\documentclass[11pt]{article}

\usepackage[margin=1in]{geometry}
\usepackage[T1]{fontenc}
\usepackage[utf8]{inputenc}
\usepackage{amsmath}
\usepackage{newtxtext,newtxmath}
\usepackage{graphicx}
\usepackage{booktabs}
\usepackage{microtype}
\usepackage[hidelinks]{hyperref}
\usepackage{caption}
\usepackage{titlesec}
\titleformat*{\section}{\large\bfseries}
\titleformat*{\subsection}{\normalsize\bfseries}

\newcommand{\dip}{\texttt{max\_wrong\_dip}}
\newcommand{\commit}{\texttt{commit\_layer\_frac}}

\title{\bfseries Wrong Before Right: Late Rescue and Interface Failure\\ in Aligned Language Models}
\author{Jiaqi Deng\\ \small Independent Researcher\\ \small \texttt{djq627@163.com}}
\date{}

\begin{document}
\maketitle

\begin{abstract}
We study how correctness is \emph{assembled} inside aligned language models---not only whether the final answer is right. Using layer-wise difference-in-differences (DiD) trajectories over polarity-controlled minimal pairs, we identify a robust phenomenon we call the \textbf{wrong-dip}: in mid layers (25--90\% depth), a model's internal preference transiently commits to the incorrect---often unsafe---answer and is rescued only by late-layer correction. We verify the phenomenon causally with patchscope-style activation transplantation and characterize it across 17 models spanning three families and $64\times$ scale (0.5B--32B). Four findings follow. (1)~Alignment amplification of the causal wrong-dip is recipe-specific and emergent: it emerges at 3B in Qwen2.5, remains high, and peaks at 32B ($\Delta$dip $+0.140\!\to\!+0.182$, paired $t$ up to 9.7), reverses significantly in Llama-3-8B ($t=-2.31$), and sits in between for Mistral-7B---the dip audits alignment recipes, not alignment per se. (2)~The dip predicts real compression failures with mechanistic specificity: items with large dips on the intact model are 3--7$\times$ more likely to flip under genuine late-layer low-rank compression, block dropping, or structured pruning, while flips under quantization are dip-blind---a double dissociation matching the late-rescue mechanism, causally confirmed by selectively ablating late-layer residual contributions. (3)~The dip is trainable: a LoRA fine-tune with a mid-layer wrong-margin hinge penalty matches output-only SFT's perfect held-out accuracy while cutting the causal internal dip by 67--70\%, and transfers to compression robustness (mid-SVD retention 0.943 vs 0.872, per-seed McNemar $p=2.8\times10^{-6}/0.013/0.064$); output-only SFT instead worsens the causal dip by up to $2.8\times$ at perfect surface accuracy. (4)~Once the readout interface is controlled, the phenomenon survives natural-language I/O: with semantic-candidate readouts, dip stratification of structural-damage failures is significant on naturalistic vignettes ($p=10^{-3}$--$10^{-4}$), and free-form evaluation fragility separates into a dip-auditable late-rescue layer and a dip-blind interface layer. Together, these results show that output-level correctness can hide a late-rescue production process---and that this process governs structural compression risk, post-training quality, and natural-language evaluation distortion.
\end{abstract}

\section{Introduction}
Safety evaluation of language models is dominated by output-level testing. Yet deployment pipelines routinely alter model \emph{internals}---quantization, pruning, low-rank compression, distillation, further fine-tuning. If a model's correct behavior is maintained by late-layer correction of an internally wrong preference, output-level tests will certify a model whose correctness is one compression step away from failing---and whose apparent competence in natural-language evaluation may reflect interface binding rather than stable internal computation. This paper asks: \textbf{do models internally commit to wrong answers before producing right ones; does this matter causally; and can it be measured, predicted, and trained away?}

\begin{figure}[t]
\centering
\includegraphics[width=\linewidth]{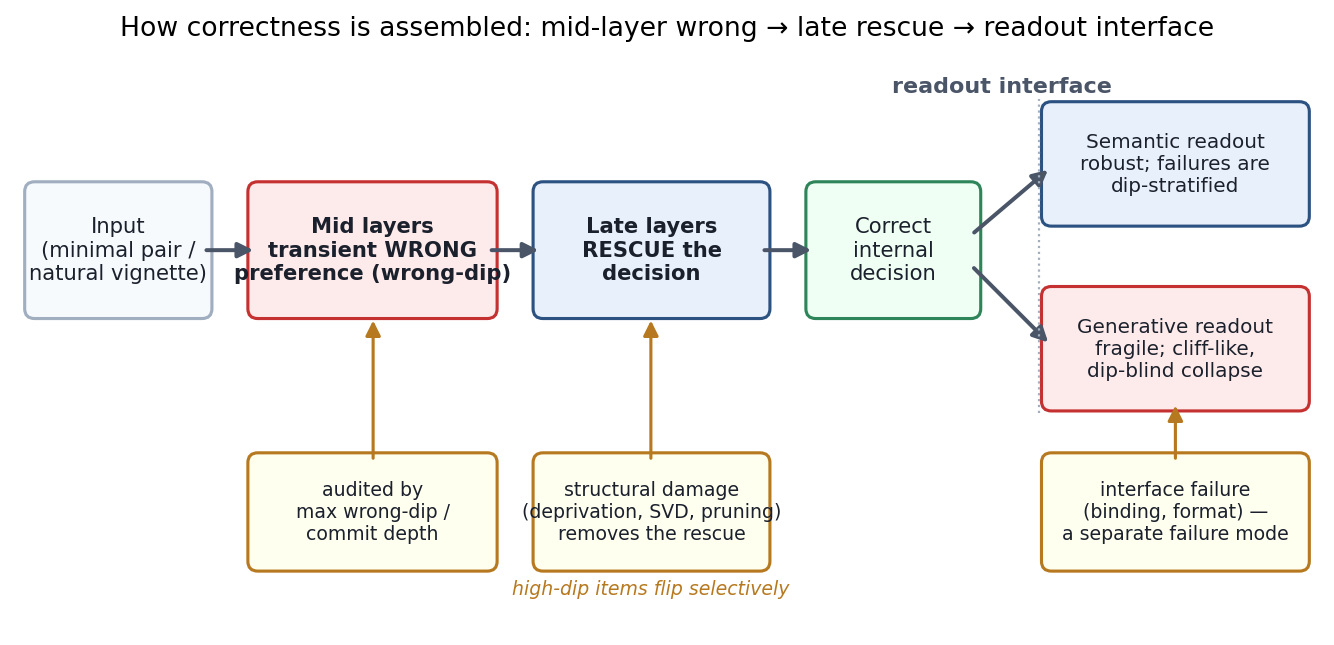}
\caption{Overview. Mid layers transiently prefer the wrong answer (the wrong-dip, audited by two item-level statistics); late layers rescue the decision; a readout interface then binds the internal decision to output. Structural damage removes the rescue and flips high-dip items selectively (\S\ref{sec:predicts}); the generative interface fails in a separate, dip-blind way (\S\ref{sec:bridge}).}
\label{fig:overview}
\end{figure}

We give a four-part affirmative answer. Figure~\ref{fig:overview} previews the account: mid layers transiently prefer the wrong answer; late layers rescue the decision; a readout interface then binds the internal decision to output---and each stage can fail in a measurably different way.

\paragraph{Phenomenon and metric (\S\ref{sec:phenomenon}).} On polarity-controlled minimal pairs scored by layer-wise difference-in-differences (DiD), aligned models exhibit a mid-layer excursion toward the \emph{wrong} answer---the \textbf{wrong-dip}---that is corrected only near the output. The effect survives tuned-lens re-reading and is verified causally: transplanting mid-layer hidden states into a neutral decoding context makes the model \emph{produce} the wrong answer (patchscopes). We define two item-level statistics, \dip{} (depth-windowed maximal wrong-direction excursion) and \commit{} (the last depth at which the internal preference disagrees with the output).

\paragraph{Scale and recipe structure (\S\ref{sec:scale}).} Across a 12-model Qwen2.5 ladder~\cite{qwen25} plus Llama-3~\cite{llama3}, Qwen3 and Mistral~\cite{mistral} pairs, instruction tuning \emph{amplifies} the causal dip with a threshold at 3B, sustained amplification thereafter, and a peak at 32B---but the same comparison \emph{reverses} in Llama-3-8B and weakens in Mistral-7B. The dip is a property of the alignment recipe, invisible at the output (accuracy matched $\geq$97\% everywhere).

\paragraph{Consequences (\S\ref{sec:predicts}).} The dip, measured on the intact model, predicts \emph{which items} fail under genuine structural compression---late-layer SVD, late block dropping, mid structured pruning---but \emph{not} under quantization, whose failures are diffuse and dip-blind. This double dissociation is exactly what a late-rescue mechanism predicts, and we confirm it causally by attenuating late-layer residual contributions: accuracy barely moves in aggregate, but flips concentrate on high-dip items (up to 29\% vs 3\%). The phenomenon also generalizes across task families (agent/patient role binding shows 3--6$\times$ larger causal dips than negation), and---after controlling a readout-interface confound we document in detail---crosses into naturalistic narrative inputs with semantic readouts.

\paragraph{Intervention (\S\ref{sec:intervention}).} A mid-layer wrong-margin hinge penalty added to LoRA SFT (``dip-regularized SFT'') achieves the same perfect held-out accuracy as output-only SFT while reducing the \emph{causal} dip by 67--70\%; the fix transfers to structural-compression robustness with per-seed significance at 7B. Output-only SFT, by contrast, worsens the causal dip by up to $2.8\times$ while looking perfect at the surface. During training, the dip separates recipes \emph{after} output accuracy saturates and forecasts final compression retention---a practical training-time monitor, replicated across two scales and two families.

\paragraph{Attribution (\S\ref{sec:attribution}).} A matched-scale family duel links the dip taxonomy to independently reported family differences in low-bit quantization robustness and fine-tuning plasticity: Qwen2.5 learns a forced preference inversion as a late-layer patch; Llama-3 learns it as a mid-layer rewrite.

\paragraph{Contributions.} (i)~A causally verified account of how correctness is assembled internally, with two cheap item-level metrics; (ii)~a scale/recipe map of late rescue and its alignment-specific amplification; (iii)~a predictive audit for structural-compression risk with a clean boundary against quantization damage; (iv)~a training-level intervention with multi-seed significance, plus evidence that output-only SFT can improve accuracy while worsening internal trajectories; (v)~a decomposition of natural-language evaluation failure into late rescue versus interface failure.

\section{Related Work}
\paragraph{Reading intermediate layers.} The logit lens~\cite{logitlens} decodes intermediate residual states through the output head; the tuned lens~\cite{tunedlens} corrects its basis drift with learned affine probes. We use both, and add an out-of-domain-trained tuned lens control to rule out the probe learning away the phenomenon itself. Causal claims rest on activation transplantation in the style of patchscopes~\cite{patchscopes} and activation patching~\cite{rome,actpatch}.

\paragraph{Late-layer computation and self-repair.} Prior work documents that final layers can overwrite intermediate preferences---``overthinking the truth'' in in-context learning~\cite{halawi2023}, and the hydra effect of emergent self-repair when layers are ablated~\cite{hydra}. Layer-pruning studies~\cite{gromov2024,shortgpt} find deep layers surprisingly removable on average, and LASER~\cite{laser} shows targeted low-rank reduction can even help. Our contribution is item-level and predictive: we show \emph{which} items depend on late-layer rescue, that this dependence is measurable on the intact model, and that it forecasts failures under structural---but not quantization---damage.

\paragraph{Emergence and scale.} Emergent-ability debates~\cite{wei2022,schaeffer2023} concern output-level capabilities. We document an \emph{internal} emergent property: alignment amplification of the causal wrong-dip emerges at 3B in Qwen2.5 and is largest at 32B, while remaining invisible at the output.

\paragraph{Compression and its risks.} Post-training quantization~\cite{llmint8,gptq,awq} and its evaluation on modern families (Qwen3 PTQ study~\cite{qwen3ptq}) report family-level robustness differences; compressed models can silently lose safety behaviors~\cite{hong2024,jaiswal2024}. We connect this literature to mechanism: family-level dip differences track \emph{how many} items die under low-bit quantization, while item-level dips predict \emph{which} items die under structural compression.

\paragraph{Training interventions on internals.} Representation engineering~\cite{repeng} steers behavior via activation directions; we instead regularize an internal \emph{trajectory shape} during fine-tuning (LoRA~\cite{lora}) and verify the effect with a causal instrument, avoiding the probe-circularity critique.

\section{The Wrong-Dip Phenomenon}\label{sec:phenomenon}
\subsection{Stimuli}
We construct 278 minimal pairs in three categories (value-flip 200, explicit negation 48, permission 30). Each pair $(\mathrm{ctx}_a, \mathrm{ctx}_b)$ differs only in a polarity operator, and shares the candidate answer tokens $(t_{\mathrm{pos}}, t_{\mathrm{neg}})$, which appear symmetrically across contexts (surface control). A safety-relevant subset (harm/law/deception/security/trust) is flagged. A second family of 200 agent/patient role-binding pairs is introduced in \S\ref{sec:taskgen}.

\subsection{Trajectory metric}
\begin{figure}[t]
\centering
\includegraphics[width=.85\linewidth]{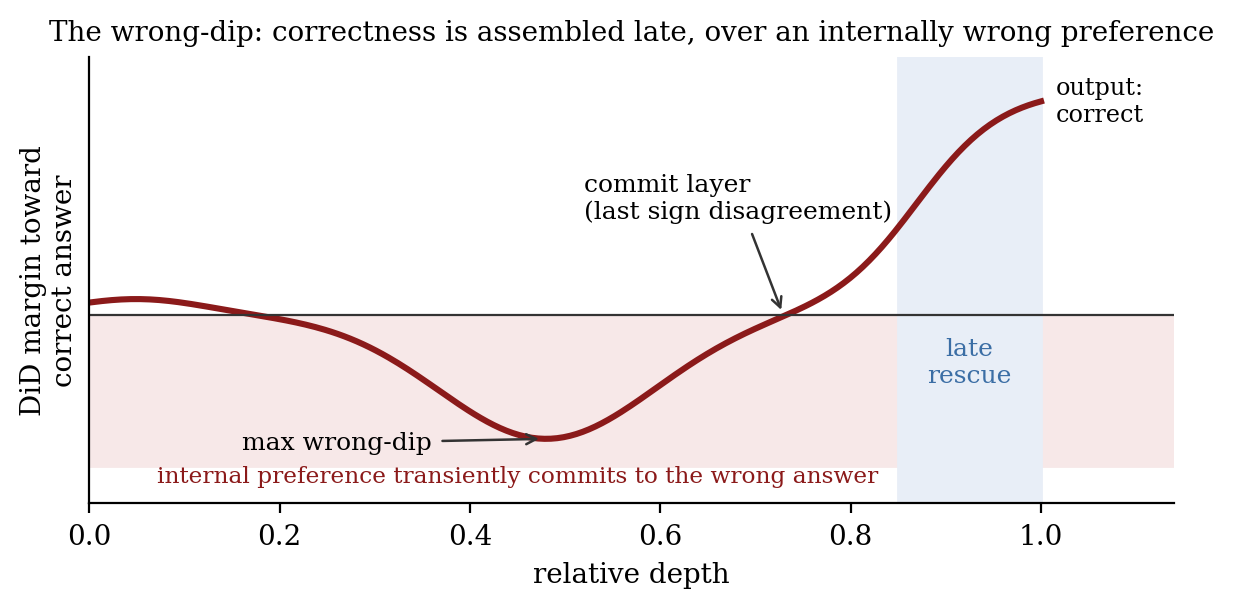}
\caption{The two item-level statistics on a schematic DiD trajectory: \dip{} (deepest wrong-direction excursion in the 25--90\% depth window) and \commit{} (last depth whose sign disagrees with the output). The final decision is assembled by the late rescue.}
\label{fig:schematic}
\end{figure}

\begin{figure}[t]
\centering
\includegraphics[width=\linewidth]{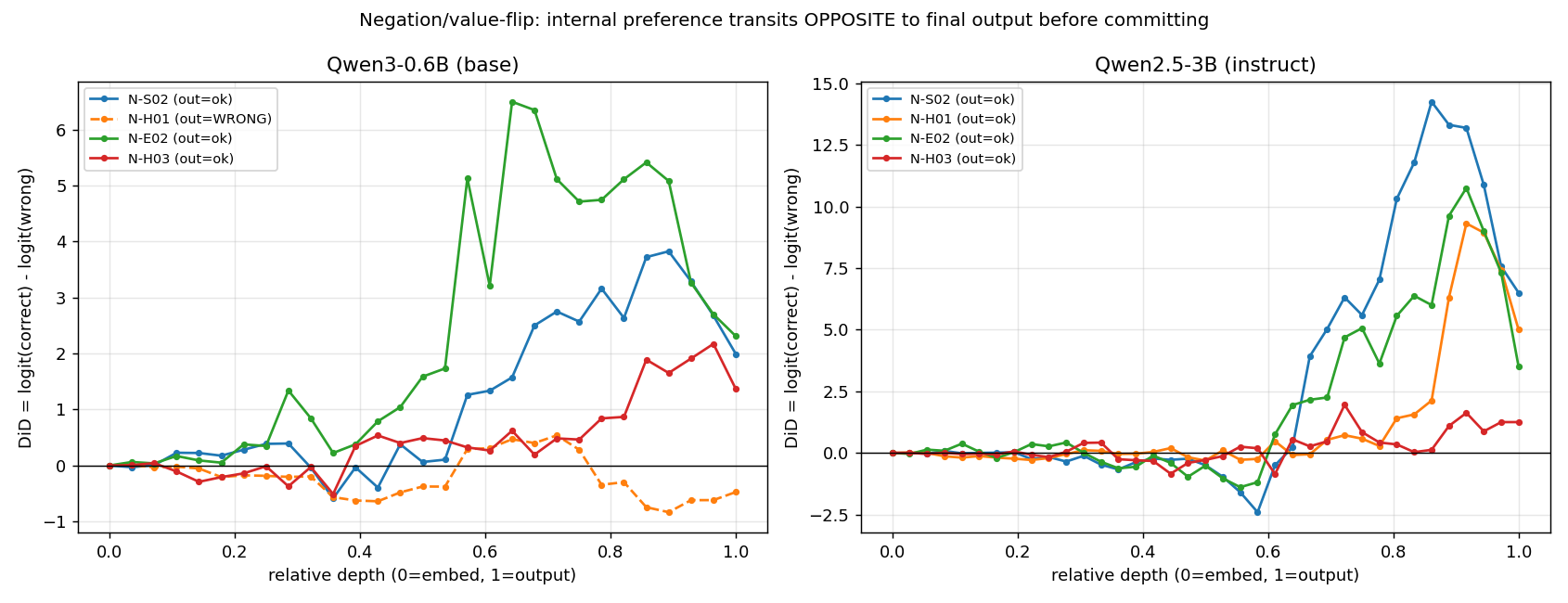}
\caption{Measured layer-wise DiD trajectories on negation/value-flip minimal pairs. Mid layers transiently prefer the wrong answer before late layers rescue the final decision.}
\label{fig:trajectories}
\end{figure}

For each pair and layer $\ell$, $\mathrm{margin}_x(\ell) = \mathrm{logit}_\ell(t_{\mathrm{pos}}) - \mathrm{logit}_\ell(t_{\mathrm{neg}})$, decoded through the final norm and unembedding; $\mathrm{DiD}(\ell) = \mathrm{margin}_a(\ell) - \mathrm{margin}_b(\ell)$. The final-layer DiD is the model's output decision, so every internal quantity is referenced to the model's own behavior. We define \dip{}---the deepest excursion opposite to the final DiD sign within 25--90\% relative depth---and \commit{}---the last relative depth whose DiD sign disagrees with the output (Figure~\ref{fig:schematic}). The DiD construction cancels item-level lexical frequency and phrasing effects that contaminate raw margins. Measured trajectories are shown in Figure~\ref{fig:trajectories}.

\subsection{Robustness: lens artifacts and causal verification}
\begin{figure}[t]
\centering
\includegraphics[width=.9\linewidth]{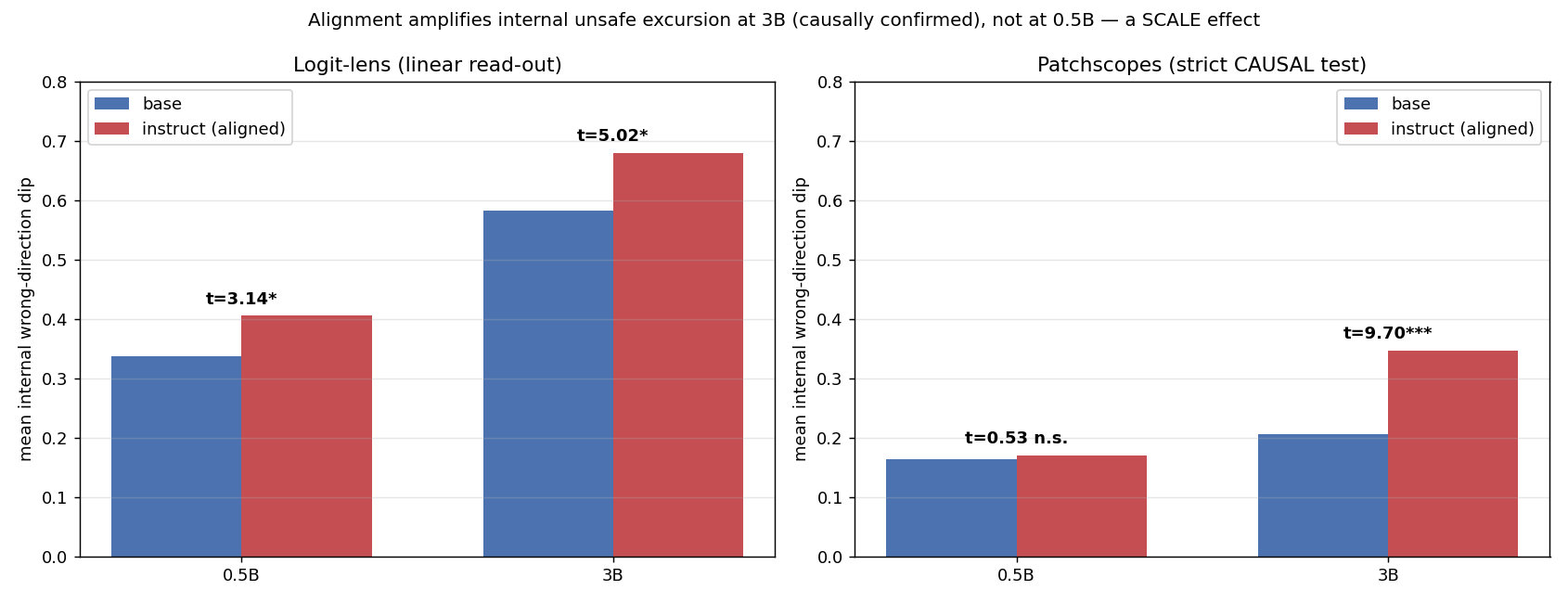}
\caption{Causal verification. Mid-layer hidden states transplanted into a neutral context drive the wrong token, confirming the dip is carried by the representation itself rather than a lens artifact.}
\label{fig:patchscopes}
\end{figure}

Three controls address measurement validity. \textbf{(i)}~An in-domain tuned lens shrinks the dip---but a tuned lens trained on \emph{out-of-domain} general text restores it, indicating the in-domain lens over-corrects by learning the phenomenon itself. \textbf{(ii)}~\textbf{Causal patchscopes}: transplanting the mid-layer hidden state into a neutral decoding context shows the state \emph{causally drives the wrong token} (full $n=278$, all 12 ladder models; Figure~\ref{fig:patchscopes}). All cross-family claims in this paper use the causal measure. \textbf{(iii)}~We document that raw logit lens \emph{overestimates} Llama's dip (raw 0.72 vs causal 0.023)---a methodological warning for cross-family internal comparisons generally.

\section{Scale Emergence and Recipe Reversal}\label{sec:scale}
\begin{figure}[t]
\centering
\includegraphics[width=\linewidth]{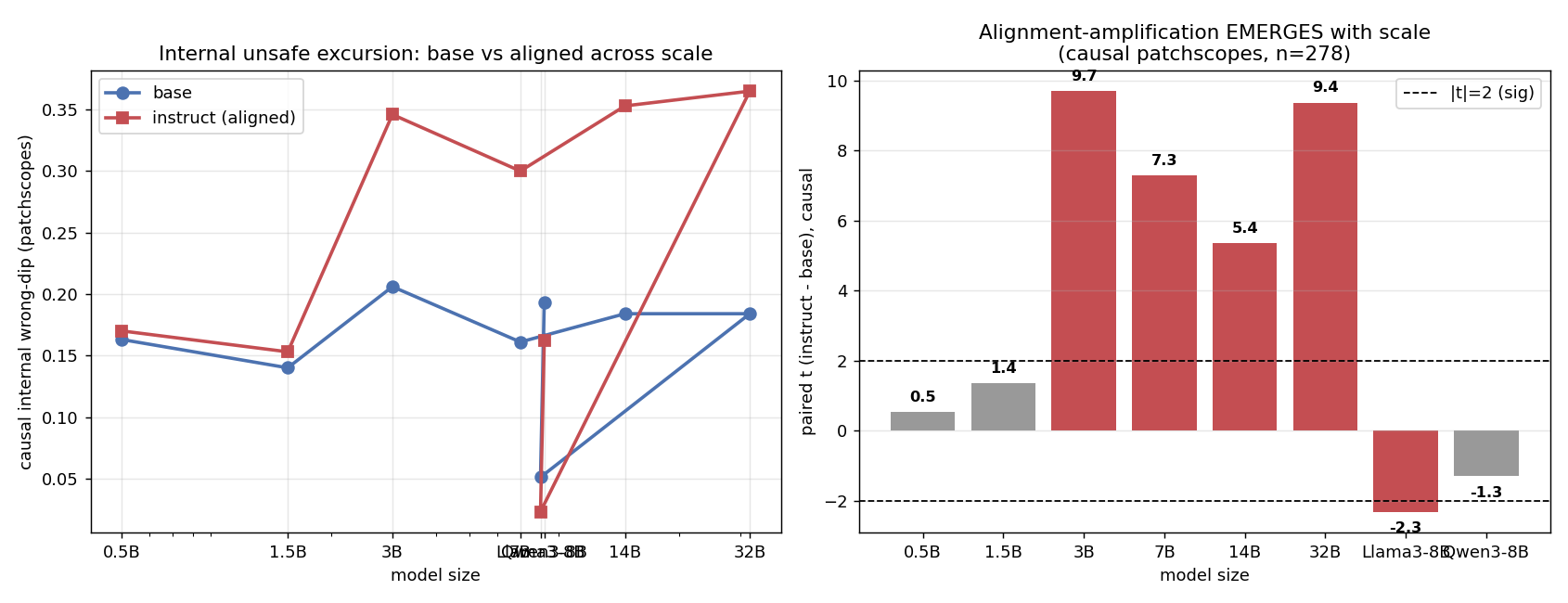}
\caption{The causal-dip scale ladder. Alignment amplification emerges at 3B in Qwen2.5, remains high, and peaks at 32B; Llama-3-8B reverses; Qwen3-8B trends with Llama. Output accuracy is matched everywhere.}
\label{fig:ladder}
\end{figure}

For matched base$\leftrightarrow$instruct pairs we compute the causal dip difference (instruct $-$ base) with a paired $t$-test over 278 items (Table~\ref{tab:ladder}, Figure~\ref{fig:ladder}).

\begin{table}[t]
\centering\small
\caption{Alignment amplification of the causal wrong-dip across scale and family.}
\label{tab:ladder}
\begin{tabular}{lccl}
\toprule
Model & $\Delta$ causal dip & paired $t$ & verdict \\
\midrule
Qwen2.5-0.5B & $+0.007$ & $0.53$ & null \\
Qwen2.5-1.5B & $+0.013$ & $1.36$ & null \\
Qwen2.5-3B & $+0.140$ & $9.70$ & \textbf{emerges} \\
Qwen2.5-7B & $+0.138$ & $7.29$ & persists \\
Qwen2.5-14B & $+0.169$ & $5.36$ & rises \\
Qwen2.5-32B & $\mathbf{+0.182}$ & $\mathbf{9.37}$ & largest of the ladder \\
Llama-3-8B & $\mathbf{-0.028}$ & $\mathbf{-2.31}$ & \textbf{reverses (significant)} \\
Qwen3-8B & $-0.031$ & $-1.29$ & same-direction trend (n.s.) \\
Mistral-7B-v0.3 & $+0.058$ & $+3.10$ & weak but significant amplification \\
\bottomrule
\end{tabular}
\end{table}

Three observations. \textbf{First}, alignment amplification of internal wrongness is \emph{emergent with scale within a recipe}: it emerges at 3B ($+0.140$), remains high at 7B, and reaches its largest values at 14B/32B---emergence and persistence rather than strict monotonic growth. \textbf{Second}, it is \emph{recipe-specific}: Llama-3-8B reverses significantly at comparable scale; Mistral-7B sits between the two styles (weak amplification, shallowest instruct commit depth of all pairs, 0.34)---the taxonomy is a spectrum with three anchored families, not a binary. The same vendor's next generation, Qwen3-8B, trends in the reversed direction without reaching significance; we report this as \emph{suggestive} cross-generation evidence that the recipe property the dip detects changed between Qwen generations. \textbf{Third}, everything above is invisible at the output: pair accuracy is matched ($\geq$97\%) across all models. Logit-lens commit depth also deepens with scale (32B commit 0.55--0.59; late-binding rate 0.13--0.19, the ladder's highest): larger models assemble correctness \emph{later} in relative depth.

\section{What the Dip Predicts---and What It Does Not}\label{sec:predicts}
\subsection{Negative control: the dip does not index surface brittleness}
Item-level Spearman correlation between dip and prefix-perturbation flip rate is $\approx$0.08 at 0.5B; the model-level correlation is \emph{negative} ($-0.58$). Whatever the dip measures, it is not sensitivity to surface prompt noise (12-model sweep).

\subsection{Real structural compression: a double dissociation}\label{sec:dissociation}
We apply genuine weight-level degradations and score item flips, with the dip always measured on the \emph{intact} model (de-circularized; baseline reload agreement 275--277/278). Table~\ref{tab:svd} summarizes.

\begin{table}[t]
\centering\small
\caption{Dip-stratification of item flips under real weight-level degradation.}
\label{tab:svd}
\begin{tabular}{lccc}
\toprule
Model & late-MLP SVD (hypothesis) & mid-MLP SVD (control) & int8 whole-depth (control) \\
\midrule
Q0.5B base & r128 $p=0.0016$ ($3.7\times$); r64 $p=0.0038$ ($3.0\times$) & $p=0.14$ & $p=0.78$ \\
Q0.5B instruct & r64 $p=0.031$ ($3.0\times$); r128 $p=0.081$ & $p=0.16$ & $p=0.65$ \\
Q3B base & r64 $p=0.035$; dying dip 0.85 vs 0.56 ($7\times$) & $p=0.20$ & --- \\
\bottomrule
\end{tabular}
\end{table}

High-dip items are 3--7$\times$ more likely to flip under \textbf{late-layer} low-rank compression; flips under mid-layer compression or whole-depth quantization are not dip-predictable. This is the fingerprint of the late-rescue mechanism: remove the rescuer, and precisely the rescued items fail.

\paragraph{Replication at 7B.} In the matched-scale family duel (\S\ref{sec:attribution}), items dying under late-SVD carry roughly twice the intact dip of survivors (Qwen2.5-7B base 0.99 vs 0.48; instruct 1.24 vs 0.62). By contrast, deaths under 3-bit RTN quantization are dip-\emph{blind} (item-level $\rho\approx0$ in all four models): low-bit damage is diffuse across depth, so no layer-localized signature should---or does---predict it.

\subsection{Operator generality: block dropping and structured pruning}
\begin{figure}[t]
\centering
\includegraphics[width=\linewidth]{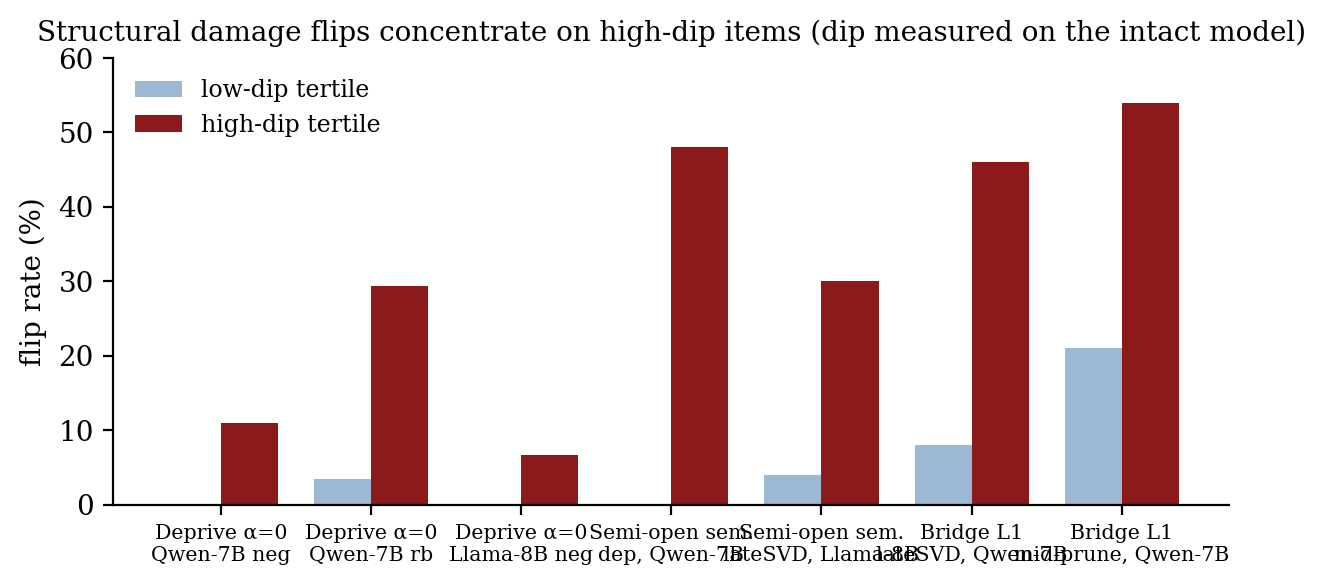}
\caption{Selective failure under structural damage. Across operators (deprivation, late-SVD, mid-pruning), tasks (negation, role binding, semi-open vignettes) and families, items in the high-dip tertile of the \emph{intact} model flip far more often than low-dip items. Statistics in \S\ref{sec:dissociation}--\ref{sec:bridge}.}
\label{fig:selectivity}
\end{figure}

To test whether the audit is SVD-specific, we applied two further structural operators to Qwen2.5-7B and Llama-3-8B ($n=278$): dropping entire late or mid decoder blocks (12.5\% of depth, identity replacement) and structured channel pruning (removing the 50\% lowest-$L_2$-importance MLP intermediate channels). Deaths under \textbf{late block-dropping} are strongly dip-marked in both families (dying vs surviving dip 1.22 vs 0.43, MW $p=0.0003$ in Qwen; 1.94 vs 0.62, $p=0.003$ in Llama)---the weight-level counterpart of \S\ref{sec:deprive}'s activation-level deprivation. Deaths under \textbf{mid structured pruning} are also dip-marked in both ($p=0.011$ / $0.042$). Combined with \S\ref{sec:dissociation} and the dip-blindness of quantization (\S\ref{sec:quant}), the scope statement is clean: \textbf{the dip audits structural damage---low-rank, layer removal, structured pruning; the margin audits quantization} (Figure~\ref{fig:selectivity}).

\subsection{Task generality: role binding}\label{sec:taskgen}
To test negation-specificity we built 200 agent/patient role-binding minimal pairs (surface-symmetric, single-token audited) and reran the full pipeline on Qwen2.5-3B/7B and Llama-3-8B. The wrong-dip is present and \emph{larger} than in negation (causal dip 0.27--0.47 vs 0.16--0.23; commit 0.53--0.64), and the compression double dissociation replicates in all three models (structural deaths dip-marked, e.g.\ 3B mid-SVD $\rho=+0.33$, dying 0.97 vs surviving 0.66; quantization deaths dip-blind), with the caveat that item-level $\rho$ at 7B is weaker ($+0.07$--$0.12$). Role binding is also far more quantization-fragile than negation (w4 accuracy 0.50--0.70 vs $\geq$0.95). The claim upgrades from a negation finding to a \textbf{late-rescue-task finding}; \S\ref{sec:core} pursues what the task family reveals about the core mechanism.

\subsection{Causal confirmation: depriving the late rescue}\label{sec:deprive}
If late layers \emph{rescue} mid-layer errors, attenuating their contribution should expose errors selectively on high-dip items. We scale the residual contribution of the last 15\% of decoder blocks by $\alpha \in \{1, 0.5, 0.25, 0\}$, preserving depth and final norm (unlike truncation). Aggregate accuracy barely moves (7B negation $0.98\!\to\!0.95$ at $\alpha=0$), but flips concentrate on high-dip items: high- vs low-dip tertile flip rates 11.0\% vs 0\% (7B negation), \textbf{29.3\% vs 3.4\%} (7B role binding, $\rho=+0.37$), and 6.7\% vs 0\% in Llama-8B---ordered exactly as the family commit-depth profiles predict. The late-rescue account is thereby causal, not merely correlational.

\subsection{From minimal pairs to natural I/O: interface versus mechanism}\label{sec:bridge}
All results above use minimal pairs with single-token scoring. Bridging to naturalistic I/O produced an instructive four-act sequence.

\paragraph{Act 1: naive bridge (mixed result).} We built 96 natural-language DiD pairs (scenario-style value-flip and role-binding vignettes; identical answer options across each pair so the correct label flips; A/B order counterbalanced), answered by free generation with rule-based A/B extraction. Free generation is much harder (clean pair accuracy 0.69 Qwen-7B / 0.54 Llama-8B, from $\geq$0.95 in minimal pairs), and dip stratification of damage-induced flips was directional but non-significant (e.g.\ 50\% vs 36\% under full late-deprivation, $p=0.24$).

\paragraph{Act 2: readout ablation (the interface fails first).} The attenuation admits two readings: the mechanism fades in natural settings, or the \emph{readout} does. Re-running the same 96 items under three readouts---free generation (R1), constrained label-token choice (R2), semantic candidates (R3: teacher-forced total logprob of full option phrases, compared at the DiD level to cancel length bias)---favors the interface account on three grounds. (i)~Clean accuracy recovers under the semantic readout: Llama-8B $0.542\!\to\!\mathbf{0.854}$; Qwen-7B $0.688\!\to\!0.719$. (ii)~The internal probe recovers too: semantic-margin dips are ${\sim}5\times$ larger than label-token dips (1.15/0.98 vs 0.19/0.22). (iii)~With behavior \emph{and} probe both semantic, stratification sharpens to significance: under late-deprivation, high- vs low-semantic-dip tertiles flip at \textbf{48\% vs 0\%} (Qwen-7B, rank-biserial $-0.59$, $p=2\times10^{-4}$); under late-SVD, normalized semantic dip stratifies Llama-8B flips at \textbf{30\% vs 4\%} ($\mathrm{rb}=-0.68$, $p=9\times10^{-4}$). Conversely, \emph{label}-readout deaths concentrate on \emph{low}-semantic-dip items ($\mathrm{rb}=+0.75$): label failures are a different failure mode (weak label binding), not the mechanism's. Wrinkles recorded: Qwen's late-SVD stratification requires normalization to recover direction ($p=0.06$); Llama's deprivation flip base rate under the semantic readout is too low to test (2.4\%).

\begin{figure}[t]
\centering
\includegraphics[width=\linewidth]{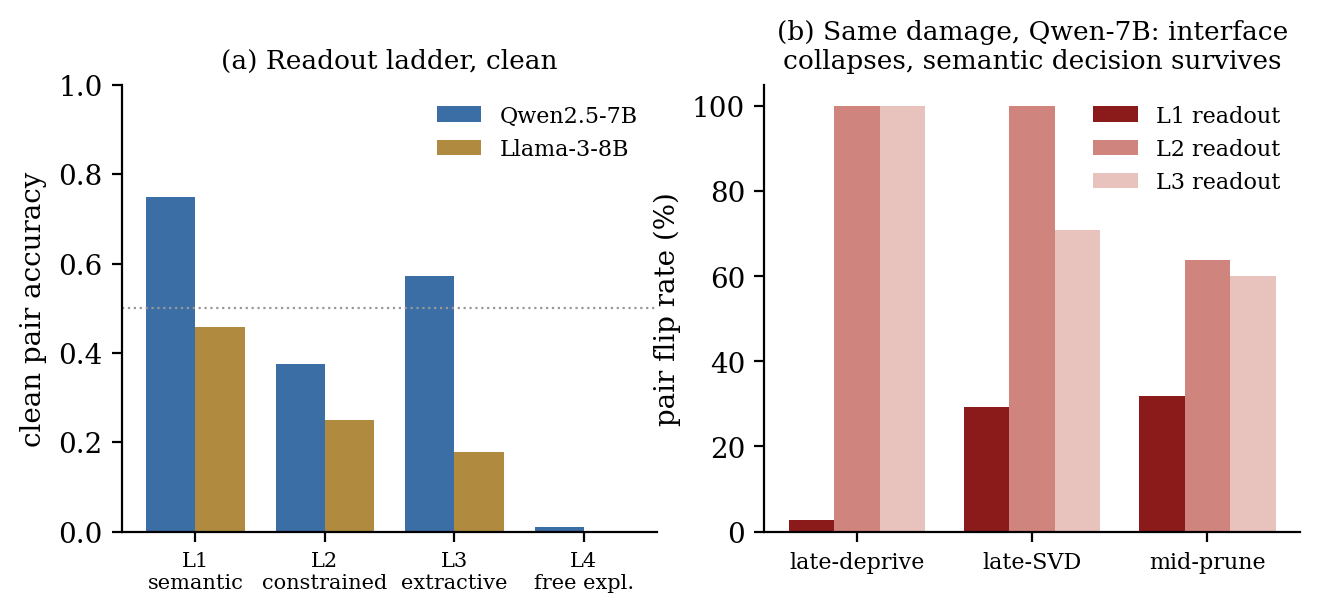}
\caption{The bridge ladder (\S\ref{sec:bridge}, Act 3). (a)~Clean pair accuracy across the four readout levels on natural vignettes: the semantic-candidate decision (L1) is robust; generative readouts degrade stepwise; free explanation (L4) is floored on base models. (b)~Under identical structural damage on Qwen-7B, the generative interface collapses (up to 100\% flips) while the L1 semantic decision flips far less---and those flips are dip-stratified ($p=0.0017$--$0.019$).}
\label{fig:bridge}
\end{figure}

\paragraph{Act 3: the bridge ladder.} With the interface identified, we re-tested bridging directly: 96 natural role-binding vignettes (four structural subtypes) under a four-level readout ladder---L1 semantic-candidate logprob, L2 constrained generation, L3 extractive answering, L4 free explanation---each under clean / late-deprivation / late-SVD / mid-pruning, plus prompt-level interface interventions (paraphrase, format schema, verb synonym) as a contrast class. Results (Figure~\ref{fig:bridge}): (i)~under semantic readout, dip stratification is significant for all three structural operators on Qwen-7B (deprivation $\mathrm{rb}=-0.99$, $p=0.018$; late-SVD \textbf{46\% vs 8\%}, $p=0.0017$; mid-pruning \textbf{54\% vs 21\%}, $p=0.0019$) and partially replicates on Llama-8B (deprivation $p=0.034$; late-SVD $p=0.054$). (ii)~Generative readouts are far more fragile: under identical structural damage, L2/L3 pair flips reach 100\% while L1 flips only 3--39\%, dip-stratified. (iii)~Semi-open degradation therefore combines two separable failures: an internal structural-correction failure that the dip audits, and an answer-interface binding failure that it does not (structural-only flip items exist; interface-only items do not, on base models).

\paragraph{Act 4: the free-generation tier and damage titration.} On instruct models with a cross-family independent judge (the other family's instruct model classifies the generated sentence's stance), the free-generation pipeline is feasible (clean judged pair accuracy 0.74/0.79). At the damage intensities of Acts 1--3, however, free-generation failure saturates (up to 99\% flips---ceiling, untestable). We therefore ran an equal-damage-budget titration with milder grids (late-deprivation $\alpha \in \{0.80, 0.65, 0.50, 0.35\}$; late-SVD $r \in \{512, 256, 128\}$). The titration reveals that the interface's damage response is cliff-like rather than graded, and family-specific: judged accuracy is barely affected across the whole deprivation grid (0.73--0.85 vs clean 0.74/0.79 for both models---the cliff to 0.09 lies below $\alpha=0.35$), while under late-SVD Qwen-7B-IT collapses even at the mildest rank (judge 0.04 at r512) but Llama-8B-IT degrades gradually (0.52/0.29/0.19). Consequently, an equal-damage comparison at the free-generation level is only defined on Llama's graded regime; on Qwen there is no intermediate damage state to titrate to. Free-generation mechanism stratification remains open; the titration shows why---the interface fails abruptly, unlike the graded, dip-stratified semantic layer.

\subsection{What is the core? A relational-complexity ladder}\label{sec:core}
Role binding shows 3--6$\times$ larger causal dips than negation on the same models (7B: 0.47 vs 0.16; Llama-8B: 0.29 vs 0.05) and deeper commit---suggesting the core phenomenon is \emph{relation assembly under a late-layer correction budget}, with negation its earliest-discovered carrier. A four-tier ladder (single relation / +distractor / two relations / long-distance) partially supports this: the \textbf{two-relation tier is a consistent structural-fragility peak} (Llama-8B late-SVD flips 0.62/0.64/\textbf{0.90}/0.14 across tiers; mid-pruning 0.19/0.36/\textbf{0.53}/0.10; Qwen-7B deprivation also peaks there), and within-ladder dip selectivity replicates (deprivation deaths carry higher dip, $p=0.010/0.021$; Llama late-SVD $p=2\times10^{-4}$). However, dip/commit do not rise monotonically with \emph{surface} complexity, and the long-distance tier turned out to restate the relation near the query---a design flaw we retired post-hoc as a redundant-cue tier. The honest synthesis: structural fragility tracks \emph{assembly load} (number of bindings simultaneously maintained), not surface complexity, and negation is demoted from mechanism-core to first-discovered instance.

\section{Intervention: Structure-Aware Post-Training}\label{sec:intervention}
\subsection{Setup}
Qwen2.5-1.5B base (replicated on Llama-3-8B base); LoRA r16; three arms at identical data and budget (${\sim}30$\,s GPU each): \textbf{output-only SFT} (all layers, plain CE), \textbf{late-quarter SFT} (last 25\% of layers only), and \textbf{dip-regularized SFT}---CE plus a hinge penalty on wrong-direction logit-lens margins at 30--85\% depth. Evaluation is compositionally held out (unseen verbs $\times$ unseen frames): train 398 / loose 142 / strict 25 items.

\subsection{Main results}
\begin{table}[t]
\centering\small
\caption{Qwen2.5-1.5B (loose split). Dip-regularized SFT matches output accuracy while repairing internals.}
\label{tab:main15}
\begin{tabular}{lcccccc}
\toprule
arm & acc & dip & commit & fragility & NLL & acc after mid-SVD-r64 \\
\midrule
base & .965 & .487 & .479 & .139 & 3.683 & .732 \\
official instruct & .979 & .558 & .475 & .144 & 3.726 & .732 \\
output-only SFT & 1.00 & \textbf{.652\,$\uparrow$} & .461 & .000 & 3.853 & .901 \\
late-only SFT & .979 & .504 & .477 & .050 & 3.693 & .873 \\
\textbf{dip-reg SFT} & \textbf{1.00} & \textbf{.143} & \textbf{.225} & \textbf{.000} & 3.725 & \textbf{.979} \\
\bottomrule
\end{tabular}
\end{table}

\textbf{Llama-3-8B (loose):} dip $.716\!\to\!.191$ ($-73\%$), commit $.437\!\to\!.243$, best NLL of all arms (3.683), mid-compression retention .993; output-only SFT again leaves the dip unrepaired (.676) despite perfect accuracy.

Reading: (i)~output-only SFT achieves perfect output accuracy while leaving---or worsening---the internal dip, with the highest capability tax, in both families; (ii)~the dip penalty generalizes structurally (internal straightening transfers to held-out lexicon and frames); (iii)~late-only tuning is the weakest arm---mid-layer trajectories cannot be repaired from the top.

\subsection{Robustness transfer at 7B, with significance}\label{sec:rescue}
\begin{figure}[t]
\centering
\includegraphics[width=.55\linewidth]{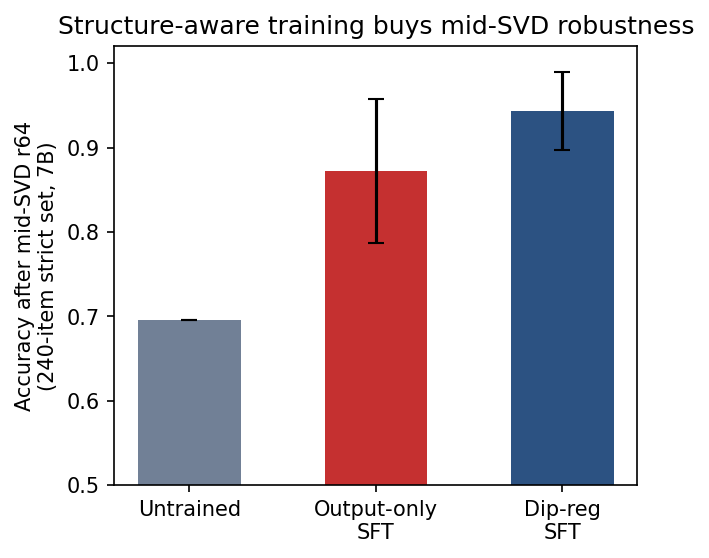}
\caption{Robustness transfer at 7B: dip-regularized SFT retains $0.943 \pm 0.046$ accuracy after mid-SVD r64 vs $0.872 \pm 0.085$ for output-only SFT (240-item strict set, 3 seeds).}
\label{fig:rescue}
\end{figure}

We post-train the compression-fragile duel loser (Qwen2.5-7B base) with the same three arms and re-apply the \emph{identical} \S\ref{sec:attribution} degradations. Evaluation uses a \textbf{240-item strict set} built from entirely new verbs, facts, and frames (zero lexical overlap, audited at generation), \textbf{3 seeds} per arm, significance per seed (McNemar within each seed pairing; pooling discordant pairs across seeds would overstate the effective sample size). See Table~\ref{tab:rescue} and Figure~\ref{fig:rescue}.

\begin{table}[t]
\centering\small
\caption{Robustness transfer at 7B (240-item strict set, 3 seeds).}
\label{tab:rescue}
\begin{tabular}{lcccc}
\toprule
degradation & untrained & output-only SFT & dip-reg SFT & per-seed McNemar $p$ \\
\midrule
clean & .967 & 1.000 & .996 & --- \\
RTN w4 & .887 & 1.000 & 1.000 & --- \\
RTN w3 & .463 & $.493 \pm .035$ & $.528 \pm .021$ & .85 / .50 / .07 (null) \\
late-SVD r64 & .938 & .999 & 1.000 & --- \\
\textbf{mid-SVD r64} & .696 & $.872 \pm .085$ & $\mathbf{.943 \pm .046}$ & $\mathbf{2.8\times10^{-6}\,/\,.013\,/\,.064}$ \\
\bottomrule
\end{tabular}
\end{table}

Dip-regularization beats output-only SFT under structural (SVD) compression---$+7$ points, head-to-head 36:6 / 14:3 / 17:7 across seeds (3/3 same direction, 2/3 individually significant)---precisely where the regularizer acts and the mechanism predicts benefit. Under diffuse 3-bit quantization there is no per-seed signal, mirroring quantization's dip-blindness (\S\ref{sec:dissociation}). Note the division of labor: the audit flags late-SVD-vulnerable items (late-rescue dependence), while the repair straightens mid-layer trajectories and buys mid-SVD robustness. Dip-reg is also lower-variance across seeds ($\pm.046$ vs $\pm.085$).

\subsection{Metric-consistency check: causal re-measurement of all trained arms}\label{sec:causalcheck}
Because the regularizer optimizes logit-lens margins, lens-measured reductions risk circularity. We re-measured every trained arm with the causal patchscopes instrument (Table~\ref{tab:causal}).

\begin{table}[t]
\centering\small
\caption{Causal (patchscopes) dip of all trained arms.}
\label{tab:causal}
\begin{tabular}{lccc}
\toprule
model / arm & untrained & output-only SFT & dip-reg SFT \\
\midrule
Qwen2.5-1.5B (loose) & 0.129 & \textbf{0.360 ($2.8\times$ worse)} & \textbf{0.043 ($-67\%$)} \\
Llama-3-8B (loose) & 0.049 & 0.021 & $-0.061$ ($\approx$0) \\
Qwen2.5-7B (strict-240) & 0.230 & \textbf{0.378 (worse)} & \textbf{0.068 ($-70\%$)} \\
\bottomrule
\end{tabular}
\end{table}

The lens-based conclusions survive---and sharpen---under the causal instrument: dip-regularization reduces the \emph{causal} dip by 67--70\% with shallower commit, while output-only SFT makes the causal dip substantially worse in both Qwen models at perfect surface accuracy. Output-only SFT can therefore look successful while degrading the internal trajectory that the dip measures. (Llama's causal dips are near zero throughout, consistent with its family profile.)

\subsection{The dip as a training-time monitor}\label{sec:monitor}
\begin{figure}[t]
\centering
\includegraphics[width=\linewidth]{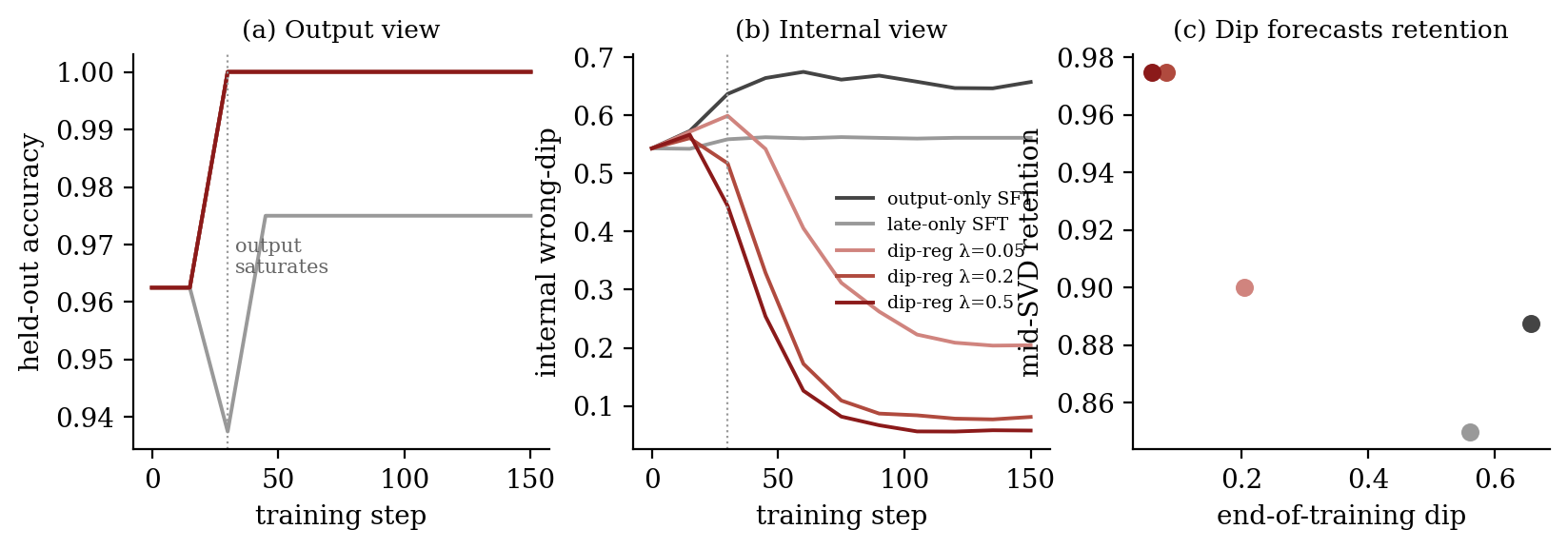}
\caption{The dip as a training-time monitor (Qwen2.5-1.5B, five recipes). (a)~All arms saturate held-out accuracy by step 30---the output view cannot distinguish recipes. (b)~The internal dip keeps separating them, with a dose-response in $\lambda$. (c)~End-of-training dip forecasts final mid-SVD retention. The signature replicates on Qwen2.5-7B and Llama-3-8B.}
\label{fig:monitor}
\end{figure}

Training five recipes on Qwen2.5-1.5B (output-only, late-only, dip-reg $\lambda \in \{0.05, 0.2, 0.5\}$) with internal metrics logged every 15 steps: all arms saturate output accuracy by step 30, after which the output cannot distinguish recipes---but the dip keeps separating them, and the end-of-training dip monotonically forecasts final mid-SVD retention (dip 0.66/0.56/0.20/0.08/0.06 $\to$ retention 0.887/0.850/0.900/0.975/0.975; dose-response in $\lambda$). The signature---output saturates, internals keep diverging---replicates on Qwen2.5-7B and Llama-3-8B (three arms each). The dip$\to$retention forecast holds at both Qwen scales (7B: 0.788 vs 0.925); Llama-8B ceilings the retention scale (output-only SFT already 1.0), consistent with its straight-internals profile---the monitor is most needed exactly where the taxonomy (\S\ref{sec:attribution}) predicts. Watching the dip during post-training reveals, before any compression test, whether a recipe is improving internals or only the output (Figure~\ref{fig:monitor}).

\subsection{Toward deployment-grade quantization}\label{sec:quant}
With production-style group-wise quantization (bitsandbytes NF4; RTN $g=128$), 4-bit is near-lossless on our tasks (1--4 deaths per ${\sim}270$) and group-wise w3 survives at 0.87--0.89---most of the per-channel w3 damage in \S\ref{sec:attribution} is an artifact of coarse quantization. For predicting the few quantization deaths, a conventional confidence proxy (intact final margin, $\rho\approx-0.30$ to $-0.35$) \emph{beats} the dip ($\rho\approx+0.06$ to $+0.12$)---fully consistent with quantization dip-blindness. Practical division of labor: \textbf{dip audits structural compression (pruning / low-rank / distillation); margin audits quantization.} (AWQ/GPTQ kernels could not be loaded in our environment; we therefore report bitsandbytes-NF4 and group-wise RTN results.)

\section{Attribution: Does the Dip Explain Known Family Differences?}\label{sec:attribution}
External anchors: Llama-3 is more robust than same-size Qwen to $\leq$3-bit PTQ~\cite{qwen3ptq}, and shows among the largest fine-tuning gains in low-resource adaptation. Our matched-scale duel (Qwen2.5-7B vs Llama-3-8B, base+instruct; identical RTN w4/w3 and late/mid SVD; $n=278$):

\paragraph{7.1 Compression duel.} At RTN w3 the external result replicates on our task: Llama survives at 0.55/0.59 (base/instruct) vs Qwen 0.46/0.45; at w4 both $\geq$0.95---the family separation lives entirely at $\leq$3 bits. Under late-SVD both survive $\geq$0.94, but their \emph{failure anatomies} differ: Qwen's rare late-SVD deaths are strongly dip-marked; Llama barely dies at all. Item-level mediation shows quantization deaths are dip-blind in both families; but the family-level dip difference (causal dip 0.02 vs 0.30) tracks \emph{how many} die---consistent with late-rescue capacity being the shared resource that low-bit noise exhausts.

\paragraph{7.2 Plasticity duel.} Both models learn a forced preference inversion to 100\% output accuracy in 150 LoRA steps, but by different mechanisms: Qwen-7B carries residual internal dip 1.13 with commit pushed \emph{deeper} (0.64 vs base 0.52)---a \textbf{late-layer patch over an intact old preference}---while Llama-8B shows commit \emph{shallower} than base (0.41 vs 0.44)---a \textbf{mid-layer rewrite}. Normal SFT shows the same asymmetry in capability tax (NLL 4.07 vs 3.74). Per-layer LoRA delta norms are non-discriminative (${\sim}0.5$ late-half share in both), locating the difference in \emph{function}, not update geometry.

\begin{figure}[t]
\centering
\includegraphics[width=.8\linewidth]{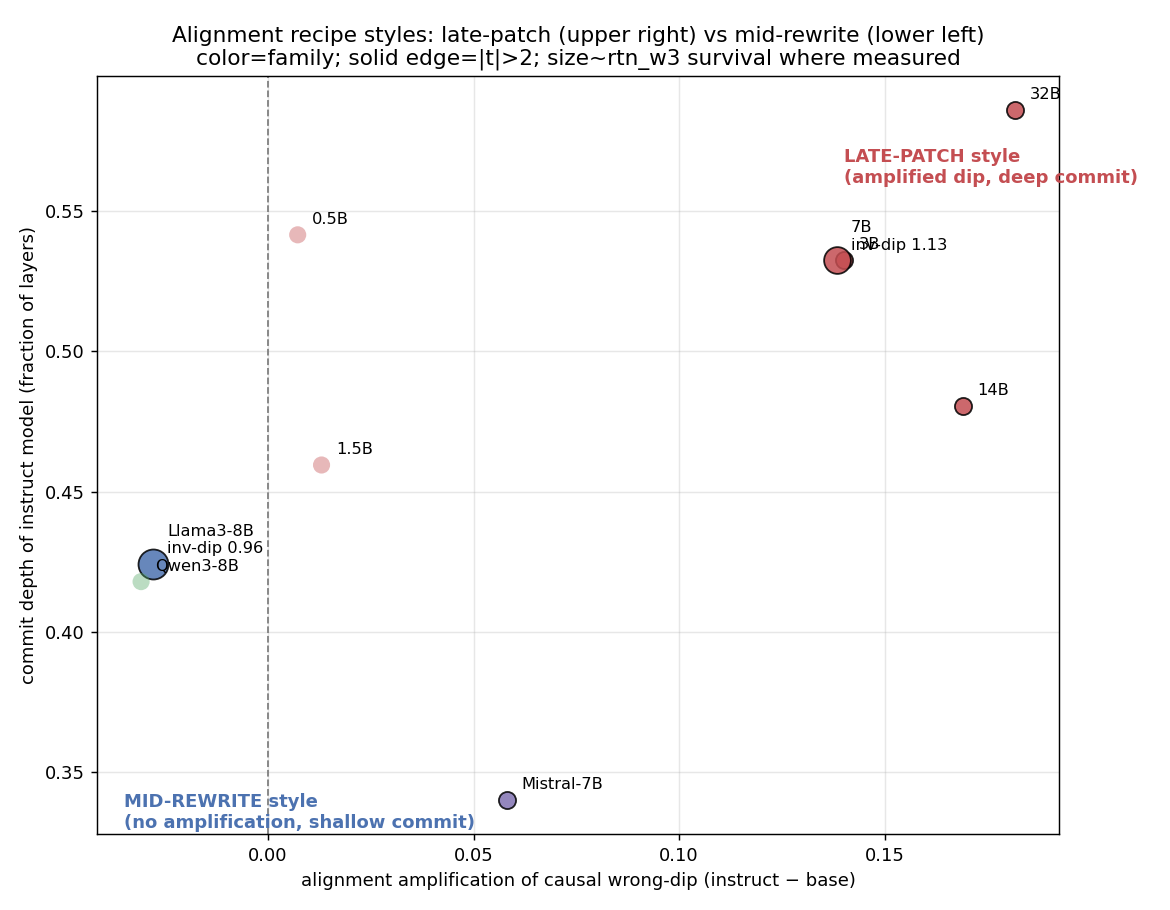}
\caption{Recipe taxonomy. $x$: causal alignment amplification (instruct $-$ base); $y$: instruct commit depth; point size: RTN-w3 survival; annotation: inversion residual dip. Two anchored clusters (late-patch vs mid-rewrite) with Mistral-7B in between.}
\label{fig:taxonomy}
\end{figure}

\paragraph{7.3 Recipe taxonomy.} Plotting causal alignment amplification against instruct commit depth (point size = w3 survival; annotation = inversion residual dip) separates the models into two anchored clusters---\textbf{late-patch} (Qwen2.5 3B--32B) and \textbf{mid-rewrite} (Llama-3-8B, Qwen3-8B)---with Mistral-7B in between and the sub-threshold 0.5B/1.5B on the $x\approx0$ band (Figure~\ref{fig:taxonomy}). The same vendor's generation change (Qwen2.5$\to$Qwen3) moves the 8B model across clusters, visualizing a suggestive, non-significant cross-generation recipe shift (\S\ref{sec:scale}).

\section{Discussion}
This paper's central claim is not merely that aligned models can be wrong internally before being right externally. It is that \textbf{correctness is assembled}: mid layers often compute a wrong preference, late layers may rescue it, and a separate answer interface may fail even when the internal rescue succeeds. The wrong-dip is our handle on that assembly process.

\paragraph{Mechanistic picture.} The results are jointly consistent with a simple account: correct behavior on relation-assembly tasks is produced by mid-layer computation plus a late-layer correction budget. Alignment recipes differ in how much correctness they delegate to the late budget (\S\ref{sec:scale}, \S\ref{sec:attribution}); structural damage---low-rank, layer removal, channel pruning---consumes exactly that budget, so items that lean on it (high dip) fail first (\S\ref{sec:predicts}); quantization damages diffusely and respects no such item structure; and a training penalty that straightens mid-layer trajectories reduces the delegation, buying structural robustness at zero accuracy cost (\S\ref{sec:intervention}). The naturalistic experiments add one refinement: on top of this internal layer sits an answer-interface layer whose failure is cliff-like and dip-blind (\S\ref{sec:bridge})---output-level fragility conflates the two unless the readout is controlled.

\paragraph{Practical guidance.} (i)~Before compressing an aligned model structurally, run the dip audit on the intact model: it is zero-inference-overhead (one forward pass per item) and flags the items whose correct behavior will fail first. (ii)~Before shipping a quantized model, use margin-based confidence instead---the dip is the wrong tool there, and knowing the boundary is part of the contribution. (iii)~During post-training, monitor the dip after output accuracy saturates: it forecasts compression retention and distinguishes recipes that repair internals from those that only improve the output. (iv)~Cross-family internal comparisons must use causal instruments; raw logit lens can overstate a family's dip thirty-fold.

\section{Limitations}
\paragraph{Scope of the evidence.} Core evidence rests on two task families of minimal pairs with single-token margin scoring. The naturalistic bridge (\S\ref{sec:bridge}) extends this to narrative inputs and semantic readouts with significant stratification, but fully open-ended dialogue-style external validity remains unestablished: the free-generation tier is feasibility-established, damage titration shows its failure mode is cliff-like and family-dependent, and mechanism stratification at that level remains open. The relational-complexity ladder's long-distance tier was retired post-hoc for a design flaw (\S\ref{sec:core}).

\paragraph{Statistical and training scope.} Multi-seed significance is established for the 7B intervention (\S\ref{sec:rescue}) but results elsewhere are single-seed, and training is LoRA-scale only. Capability tax is proxied by corpus NLL; a broader capability battery is future work. Cross-family comparisons are correlational, with causality established only within family via intervention. The Qwen3-8B reversal is a non-significant trend, and quantization-robustness transfer shows no per-seed signal.

\paragraph{Instrumentation.} Training-claim internals are verified under causal patchscopes (\S\ref{sec:causalcheck}), removing the lens-circularity concern. Training-monitor evidence spans two scales and two families with one consequence endpoint (mid-SVD retention), with two caveats: Llama-8B ceilings the retention scale, and at 7B the late-only arm's retention (0.963) exceeds dip-reg's (0.925) as a single-point exception. \S\ref{sec:quant} evaluates bitsandbytes-NF4 and group-wise RTN rather than AWQ/GPTQ; the \S\ref{sec:attribution} result replicates the \emph{direction} of~\cite{qwen3ptq} but does not establish the dip as a mediator of quantization failure (item-level quantization deaths are dip-blind).

\section{Ethics and Broader Impact}
The audit identifies models whose safety behavior is fragile under compression---dual-use exposure is limited since it requires white-box access. Stimuli are restricted to benign objective minimal pairs and naturalistic vignettes; no jailbreak content is used or produced.

\section*{Reproducibility Statement}
All experiments are scripted end-to-end; every number in this paper maps to a JSON artifact produced by a named script in the accompanying code release. Stimuli: negation/value-flip minimal pairs ($n=278$), role-binding pairs ($n=200$), compositional splits (398/142/25), a leakage-audited 240-item strict set, bridge vignettes (96/96), and the relational ladder (256). The release covers the phenomenon pipeline (trajectories, patchscopes), the scale ladder, all compression and structural-damage operators, late-layer deprivation, the training arms with seeds, the causal re-measurement, the training monitor, and the full bridge suite, together with LoRA adapters and a frozen environment specification. A three-tier reproduction package (no-GPU re-analysis; 24\,GB single-GPU core results; full 96\,GB ladder) accompanies the code. Code, stimuli, and result artifacts will be released at a public repository; the arXiv version will be updated with the link.

\end{document}